\documentclass[letterpaper]{article} 
\usepackage{aaai2026}  
\usepackage{times}  
\usepackage{helvet}  
\usepackage{courier}  
\usepackage[hyphens]{url}  
\usepackage{graphicx} 
\usepackage{natbib}  
\usepackage{caption} 
\usepackage{algorithm}
\usepackage{algorithmic}

\usepackage{amsmath}
\usepackage{amsfonts}
\usepackage{amssymb}
\usepackage{mathtools}
\usepackage{ragged2e}
\usepackage{caption}
\usepackage{subcaption}
\usepackage{booktabs}
\usepackage{multirow}
\usepackage{makecell}
\usepackage{colortbl}
\usepackage{float}
\usepackage{array}
\usepackage{diagbox}
\usepackage{tabularx}   
\usepackage{siunitx}    
\newcommand\ChangeRT[1]{\noalign{\hrule height #1}}
\usepackage{newfloat}
\usepackage{listings}
\DeclareCaptionStyle{ruled}{labelfont=normalfont,labelsep=colon,strut=off} 
\floatstyle{ruled}
\newfloat{listing}{tb}{lst}{}
\floatname{listing}{Listing}
\title{SRSplat: Feed-Forward Super-Resolution Gaussian Splatting from \\ Sparse Multi-View Images}
\author{
    Xinyuan Hu\textsuperscript{\rm 1} \equalcontrib,
    Changyue Shi\textsuperscript{\rm 1} \textsuperscript{\rm 2} \equalcontrib,
    Chuxiao Yang\textsuperscript{\rm 1},
    Minghao Chen\textsuperscript{\rm 1},
    Jiajun Ding\textsuperscript{\rm 1} \thanks{Corresponding author.}, \\
    Tao Wei\textsuperscript{\rm 2} \textsuperscript{\rm 3},
    Chen Wei\textsuperscript{\rm 3},
    Zhou Yu\textsuperscript{\rm 1},
    Min Tan\textsuperscript{\rm 1}
}
\affiliations{
    \textsuperscript{\rm 1}School of Computer Science and Technology, Hangzhou Dianzi University, \\
    \textsuperscript{\rm 2}School of AI for Science, Peking University, \\
    \textsuperscript{\rm 3}Li Auto Inc. \\

    \{xinyuan, shicy, chuxiao\_yang, djj\}@hdu.edu.cn
}

\usepackage{bibentry}

\begin{document}

\maketitle

\begin{abstract}
Feed-forward 3D reconstruction from sparse, low-resolution (LR) images is a crucial capability for real-world applications, such as autonomous driving and embodied AI. However, existing methods often fail to recover fine texture details. This limitation stems from the inherent lack of high-frequency information in LR inputs. To address this, we propose \textbf{SRSplat}, a feed-forward framework that reconstructs high-resolution 3D scenes from only a few LR views. Our main insight is to compensate for the deficiency of texture information by jointly leveraging external high-quality reference images and internal texture cues. We first construct a scene-specific reference gallery, generated for each scene using Multimodal Large Language Models (MLLMs) and diffusion models. To integrate this external information, we introduce the \textit{Reference-Guided Feature Enhancement (RGFE)} module, which aligns and fuses features from the LR input images and their reference twin image. Subsequently, we train a decoder to predict the Gaussian primitives using the multi-view fused feature obtained from \textit{RGFE}. To further refine predicted Gaussian primitives, we introduce \textit{Texture-Aware Density Control (TADC)}, which adaptively adjusts Gaussian density based on the internal texture richness of the LR inputs. Extensive experiments demonstrate that our SRSplat outperforms existing methods on various datasets, including RealEstate10K, ACID, and DTU, and exhibits strong cross-dataset and cross-resolution generalization capabilities. 
\end{abstract}

\begin{links}
    \link{Project Page}{https://xinyuanhu66.github.io/SRSplat/}
\end{links}

\begin{figure*}[t]
    \centering
    \includegraphics[width=\textwidth]{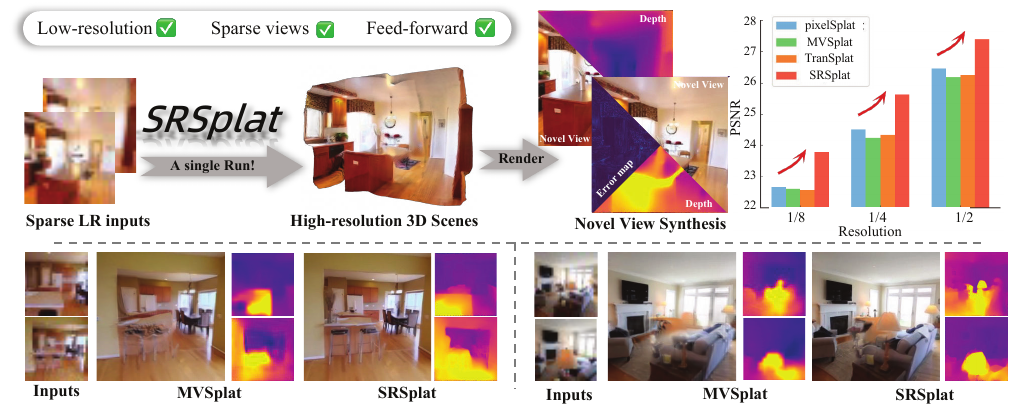}
    \caption{
    In this work, we propose SRSplat, a novel feed-forward framework that reconstructs high-quality 3D scenes with only sparse and LR input views. SRSplat demonstrates superior performance and is capable of handling low-resolution, sparse-view inputs and real-time reconstruction, thereby offering greater functionality and practicality in realistic applications.}
    \label{fig:dive}
\end{figure*}

\section{Introduction}

Reconstructing 3D scenes from 2D images is a fundamental task in computer vision and computer graphics~\cite{mildenhall2021nerf,chen2022tensorf,kerbl20233d, shi2025mmgs,shi2025realm,muller2022instant}. It plays a crucial role in various applications, such as embodied AI~\cite{huang2023visual} and autonomous driving~\cite{tian2025drivingforward}. However, in these real-world applications, traditional 3D reconstruction methods like NeRF~\cite{mildenhall2021nerf} or 3DGS~\cite{kerbl20233d} face several key challenges: 1) \textbf{\textit{Resolution Constraints:}} Due to hardware and sensor limitations, acquiring sufficiently high-resolution (HR) images for accurate 3D reconstruction is often impractical.
2) \textbf{\textit{Few-Shot Challenges:}} In practice, acquiring high-quality dense views is prohibitively costly and impractical.
3) \textbf{\textit{Real-Time Reconstruction:}} Fields like robotics require real-time 3D reconstruction, necessitating efficient feed-forward algorithms. 

Our goal is to build a feed-forward 3D reconstruction framework that simultaneously tackles the three practical hurdles so that high-quality 3D scenes can be reconstructed from only limited LR images. The recent emergence of feed-forward 3DGS~\cite{charatan2024pixelsplat,chen2024mvsplat,zhang2025transplat,tang2024hisplat,chen2024mvsplat360,fei2024pixelgaussian} methods has revolutionized 3D scene reconstruction through their real-time reconstruction capabilities. These methods leverage feed-forward networks to make direct 3D Gaussian predictions, eliminating the need for per-scene optimization. 
However, when provided with low-resolution (LR) input images, existing methods struggle to reconstruct high-quality scenes and often exhibit a loss of texture details. This limitation stems from the inherent lack of high-frequency texture information in LR images compared to their HR counterparts~\cite{2024srgs}. 

To this end, we propose \textbf{SRSplat}, a novel feed-forward framework that reconstructs HR 3D scenes from sparse LR views. Our main insight is to compensate for the deficiency of texture information by jointly leveraging external high-quality reference images and internal texture cues.
Inspired by previous reference-based 2D image super-resolution (SR) methods~\cite{lu2021masa,sun2024coser,yang2020learning,bosiger2024mariner,jiang2021robust,cao2022reference}, we first construct a scene-specific \textit{Reference Gallery}. In this gallery, each reference image is generated as a twin image of its corresponding input scene. Specifically, given a scene, we first employ Multimodal Large Language Models (MLLMs)~\cite{achiam2023gpt} to generate a concise semantic description. The resulting description is then used to prompt a pre-trained diffusion model~\cite{labs2025flux1kontextflowmatching} to synthesize high-quality reference twins. Our novel reference twin generation technique synthesizes images that mirror the target scene’s characteristics but in a HR way, supplying more high-frequency cues for the subsequent process.

With the \textit{Reference Gallery}, we propose the \textit{Reference-Guided Feature Enhancement (RGFE)} module to integrate this external information. For the input images and their corresponding HR reference twins, we first use a shared CNN for multi-scale feature extraction. Then \textit{RGFE} performs coarse-to-fine correspondence matching for the extracted features. Finally, a fuse network is used to map the distribution of reference features to LR features, thereby transferring high-frequency information from the reference twins. 
The resulting fused features are then fed to a decoder to predict the Gaussian primitives. However, since the proposed Gaussian prediction decoder predicts a single Gaussian primitive for each pixel, texture-rich regions are difficult to optimize. To further refine predicted Gaussian primitives, we propose \textit{Texture-Aware Density Control (TADC)}. Specifically, \textit{TADC} builds a learnable texture richness perceptron that perceives texture richness from LR inputs. Based on the internal texture richness, Gaussian primitives can adaptively adjust the density in the scene. 

Our contributions can be summarized as follows:
\begin{itemize}

    \item  We propose SRSplat, a feed-forward framework that generates HR 3D scenes from sparse LR 2D images. To the best of our knowledge, this is the first work to solve this task in a feed-forward manner.

    \item  We construct a scene-specific reference twin gallery using MLLMs and diffusion priors. We propose the \textit{RGFE} and \textit{TADC} module to compensate for the deficiency of texture information by leveraging external reference image and internal texture richness map.

    \item  Experimental results on various public datasets demonstrate that SRSplat outperforms existing methods and exhibits strong cross-dataset and cross-resolution generalization capabilities.

\end{itemize}

\begin{figure*}[t]
    \centering
    \includegraphics[width=\linewidth]{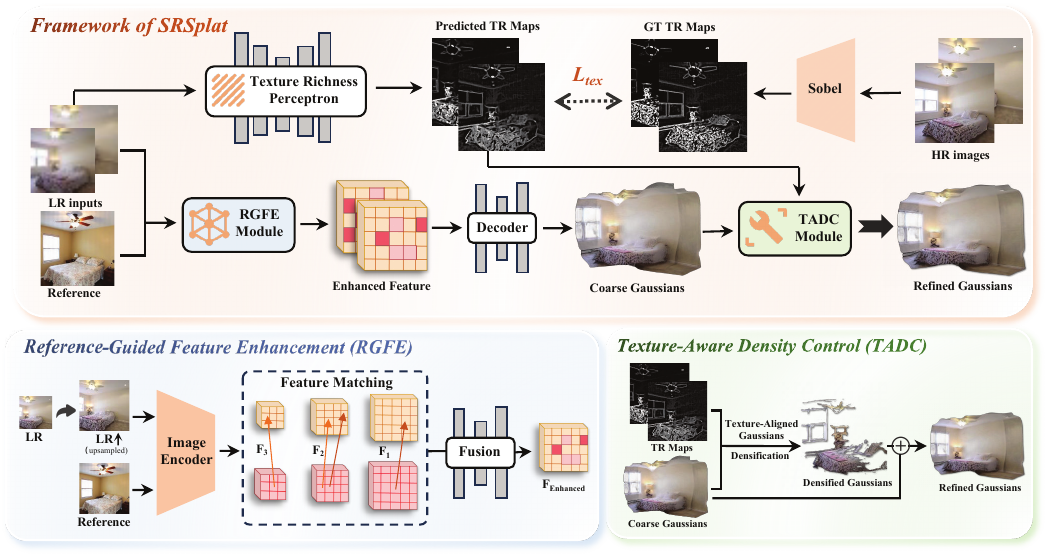}
    \caption{Framework of SRSplat. Our method takes LR images and their corresponding reference as inputs. The \textit{RGFE} module first extracts multi-scale features and effectively fuses these features. Upon decoding the Gaussian primitives, \textit{TADC} adjusts Gaussian density adaptively according to the richness of texture generated by a texture richness perceptron.} 
    \label{fig:pipeline}
\end{figure*}

\section{Related Works}
\noindent \textbf{Feed-Forward 3D Reconstruction.}  Recently, 3DGS~\cite{kerbl20233d}-based feed-forward reconstruction has emerged as a key approach for efficient 3D scene representation and novel view synthesis. PixelSplat~\cite{charatan2024pixelsplat} introduces a polar line-based Transformer architecture to model cross-view correspondence and predict depth distribution. MVSplat~\cite{chen2024mvsplat} constructs cost volumes via plane sweeping to achieve enhanced geometric reconstruction accuracy. TranSplat~\cite{zhang2025transplat} adopts a Transformer-based architecture, combining monocular depth priors to refine the sparse-view reconstruction results. DepthSplat~\cite{xu2025depthsplat} presents a robust multi-view depth model by leveraging pre-trained monocular depth features~\cite{yang2024depth}, thereby enabling high-quality feed-forward reconstructions. Despite these advancements, existing methods often fail to capture high-frequency details from low-resolution inputs, leading to artifacts and reduced quality. To address these limitations, we propose SRSplat, a novel framework that enables super-resolution reconstruction with low-resolution inputs.

\noindent \textbf{Super-Resolution Novel View Synthesis.} Super-resolution novel view synthesis aims to reconstruct high-resolution 3D scenes from only low-resolution multi-view inputs. As a pioneer in the field, NeRF-SR~\cite{wang2021nerf-sr} leverages the sub-pixel constraint to optimize a HR scenes. SRGS~\cite{2024srgs} is the first framework to synthesize HR novel views based on 3DGS using the SwinIR model~\cite{liang2021swinir}, but suffers from multi-view inconsistencies. Recently, S2Gaussian~\cite{wan2025s2gaussian} employs a multi-stage training paradigm to reconstruct high-resolution 3D scenes solely from only sparse and low-resolution input views. However, per-scene optimization methods rely on iterative optimization to achieve the final 3D representation and exhibit poor generalization ability. In contrast, feed-forward inference methods reconstruct the entire scene in a single feed-forward pass, demonstrating strong potential in super-resolution novel-view synthesis.

\noindent \textbf{Reference-Based Super-Resolution.}
Reference‐based SR methods aim to enhance low‐resolution inputs by leveraging high‐frequency details from high‐quality reference images. MASA‐SR~\cite{lu2021masa} introduces attention mechanisms to adaptively fuse reference textures into LR feature representations. More recently, RefSR-NeRF~\cite{huang2023refsr} further extends this idea to 3D by integrating reference images as auxiliary information. Yet, in real-world scenarios, it is often impractical to assume that every scene can be paired with a corresponding reference image. CoSeR~\cite{sun2024coser} leverages prior knowledge from large-scale text-to-image diffusion models to synthesize high-quality reference images for super-resolution. This motivates the development of an automated reference-generation approach.

\begin{figure*}
    \centering
    \includegraphics[width=0.98\linewidth]{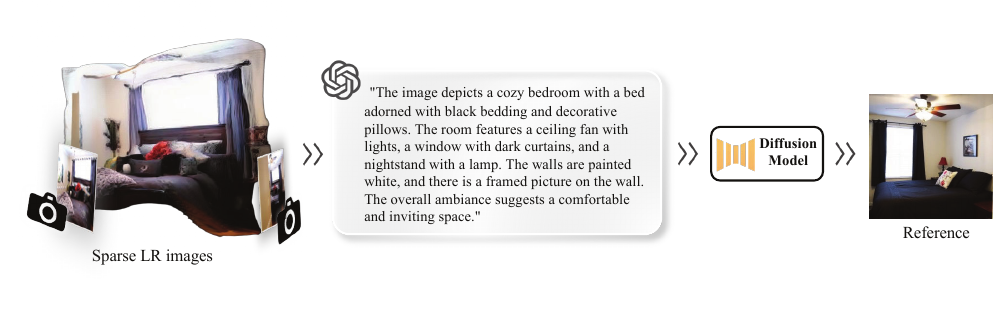}
    \caption{Pipeline of reference gallery generation. Given LR input images for each scene, the MLLM produces semantic descriptions. Subsequently, the diffusion model uses these descriptions to generate reference images tailored to the scene.    
    } 
    \label{fig:refgpt}
\end{figure*}

\begin{figure}[t]
    \centering
    \includegraphics[width=\linewidth]{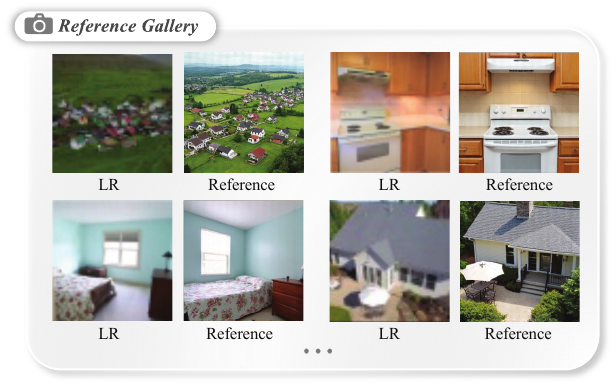}
    \caption{Reference gallery examples. Reference twin images share details similar to the LR images.
    } 
    \label{fig:refgal}
\end{figure}

\section{Framework of SRSplat}
The overall framework of SRSplat is illustrated in Fig.~\ref{fig:pipeline}. 
We first establish a reference gallery in Sec.~\ref {sec:ref_gallery}.
Then we employ \textit{RGFE} to fuse features derived from input and reference images in Sec.~\ref{sec:RGFE}. Based on the fused features, we train a decoder to predict Gaussian attributes in Sec.~\ref {depth estimation}. Finally, we use \textit{TADC} to adaptively adjust the density of Gaussian primitives according to texture intensity in Sec.~\ref {sec:TADC}.

\subsection{Reference Gallery Preparation}
\label{sec:ref_gallery}
Drawing inspiration from 2D reference-based SR methods~\cite{lu2021masa,bosiger2024mariner}, we aim to mitigate the high-frequency loss of LR input multi-view images with the information from external reference images. To achieve this, we first establish a reference gallery. Each image in this gallery serves as a twin of each 3D scene, mirroring its characteristic layouts. The process is illustrated in Fig.~\ref{fig:refgpt}. 

We first leverage GPT-4o to generate semantic descriptions of input LR multi-view images. 
For a set of LR input images (downsampled by a factor of $P$ from training set)
 of a specific scene $\{\,I^\text{LR}_i\}_{i=1}^{N}, 
I_i \in \mathbb{R}^{\frac{H}{P} \times \frac{W}{P} \times 3}$,
we employ MLLM to directly process the LR image set and generate semantic descriptions $ \mathcal{P} = {\text{MLLM}}(\{\,I^\text{LR}_i\}_{i=1}^{N})$.
The generated descriptions $\mathcal{P}$ capture the overall layout of the scene, key objects and their relationships, providing clear semantic guidance for the subsequent process.
With the semantic descriptions, we utilize a 2D diffusion model~\cite{labs2025flux1kontextflowmatching} $\mathcal{D}$ to generate a HR reference twin $r = \mathcal{D}(\mathcal{P}), r \in \mathbb{R}^{H \times W \times 3}$. By leveraging strong priors of MLLM and diffusion model, the reference twin faithfully preserves scene's layout and remains semantically aligned with the original inputs (as demonstrated in Fig.~\ref{fig:refgal}).

\subsection{Reference-Guided Feature Enhancement}
\label{sec:RGFE}
\noindent \textbf{\textit{Multi-Scale Feature Extraction.}} For an LR image $I^\text{LR} \in\mathbb{R}^{\frac{H}{P}\times \frac{W}{P}\times 3}$, we employ a shared CNN encoder~\cite{xu2022gmflow,xu2023unifying} $\mathcal{E}$ to extract multiscale features from both the upsampled input images $I\bigr\uparrow\in\mathbb{R}^{H\times W\times 3}$ and its corresponding reference $r\in\mathbb{R}^{H\times W\times 3}$. In practice, we employ three levels, each halving the resolution of the previous one. The encoded outputs are
\begin{equation}
\label{eq:feature}
\left\{
\begin{aligned}
\{F_{l}^{\mathrm{I}}\}  &= \mathcal{E}(I\bigr\uparrow),\\
\{F_{l}^{\mathrm{ref}}\} &= \mathcal{E}(r),
\end{aligned}
\right.\quad l = 1,2,3,
\end{equation}
where $F_{l}^{\mathrm{I}},\,F_{l}^{\mathrm{ref}}\in\mathbb{R}^{H_{l}\times W_{l}\times C^{{feature}}}$ denote the features at scale $l$ with resolution $H_{l}\times W_{l}$ and channel $C^{feature}$.

\noindent \textbf{\textit{Feature Matching and Fusion.}} Following previous researches~\cite{lu2021masa,bosiger2024mariner}, SRSplat matchs the features between input and reference features using cosine similarity. Specifically, it first performs coarse-to-fine matching, starting with a coarse grid using a stride, followed by dense matching within a fixed-size window around the initial correspondences. This procedure yields a mapping \(m\) from input feature indices to corresponding reference feature indices:
\begin{equation}
\label{eq:masa}
    m_{I\to ref}^l : (x,y)\in{F}_{l}^{I} \;\longmapsto\; \bigl\{(u,v)\in{F}_{l}^{ref},\;s\in\mathbb{R}\bigr\},
\end{equation}
where $(u,v)$ is the best match coordinate in the reference for the input coordinate $(x,y)$ and $s$ is the corresponding match score. Next, we warp the reference features according to \(m_{I\to ref}^l\) and weight them by their match scores, yielding the warped reference features \(\{F_1^{\mathrm{ref}\to\mathrm{I}}, F_2^{\mathrm{ref}\to\mathrm{I}}, F_3^{\mathrm{ref}\to\mathrm{I}}\}\). Weighting the warped features by their matching scores reduces low-confidence correspondences. This enables the model to use the reference features only when they have a confident match.
Finally, we fuse the warped reference features \(F_{l}^{\mathrm{ref}\!\rightarrow\!\mathrm{I}}\) with the image features \(F_{l}^{\mathrm{I}}\) using a learnable fusion network $\mathcal{H}$:
\begin{equation}
\label{eq:enhanced}
F_{enhanced} = \mathcal{H}\bigl( \{F_l^I\},\,\{F_l^{ref\to I}\}\bigr),
\end{equation}
where $F_{enhanced}\in\mathbb{R}^{\frac{H}{4}\times \frac{W}{4}\times C^{enhanced}}$ represents the final enhanced feature. To further enhance information exchange across views, we feed the features into the Swin Transformer~\cite{liu2021swin}, which employs both cross-attention and self-attention mechanisms.

\subsection{Depth Estimation and Gaussian Prediction}
\label{depth estimation}

We utilize the cost volume matching in Multi-View Stereo (MVS)~\cite{yao2018mvsnet,cao2022mvsformer} for depth estimation. Specifically, we first construct a depth candidate \(\displaystyle d^i_{\mathrm{cand}}\in\mathbb{R}^{\frac{H}{4}\times \frac{W}{4}\times D}\)
using the plane-sweep stereo approach~\cite{yao2018mvsnet}. Then, we warp the feature of the j-th view \(F^j\) into the i-th view via the two camera projection matrices \(P^i\) and \(P^j\in\mathbb{R}^{4\times4}\), producing
\begin{equation}
\label{eq:warp}
F^{ij}_{\mathrm{Warp}}
=\mathrm{Warp}\bigl(F^j,\;P^i,\;P^j,\;d^i_{\mathrm{cand}}\bigr),
\end{equation}
where Warp denotes the warping operation~\cite{xu2023unifying} and D denotes the depth dimension.

We then obtain the cost volume by computing the dot product between \(F^i\) and $F^{ij}_{\mathrm{Warp}}$:
\begin{equation}
C^i \;=\; \frac{F^i \,\otimes\, F^{ij}_{\mathrm{Warp}}}{\sqrt{C}},
\label{eq:cost_volume}
\end{equation}
where \(\otimes\) denotes element‐wise multiplication followed by summation over the channel dimension and a CNN-based upsampler.  

Finally, denoting the vector of depth candidates by 
\(\,G=[d_1,d_2,\dots,d_D]\in\mathbb R^D\), we obtain the depth map:
\begin{equation}
\label{eq:depth}
D^i \;=\; \operatorname{softmax}\bigl(C^i\bigr)\,G,\quad
D^i \in \mathbb{R}^{H\times W}.
\end{equation}

We follow previous research~\cite{chen2024mvsplat,charatan2024pixelsplat} to predict Gaussian parameters \(G_{\text{coarse}}\), including the Gaussian center \(\mu_i\), opacity $\alpha_i$, covariance $\Sigma_i$, and color $c_i$. Specifically, we utilize the depth map $D^i$ to unproject the 2D pixel to the 3D space location as the Gaussian center and predict opacity using a simple MLP layer. We calculate color from the predicted spherical harmonic coefficients and use a scaling matrix $s$ and a rotation matrix $R(\theta)$ to represent the covariance matrix.
\begin{equation}
    \Sigma_i = R(\theta)^\top \,\mathrm{diag}(s)\,R(\theta).
\end{equation}

\begin{figure}[t]
    \centering
    \includegraphics[width=\linewidth]{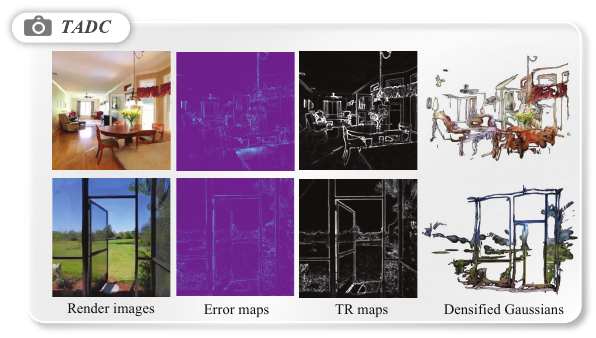}
    \caption{Error maps show the intensity of inconsistency between the rendered image and the ground truth. We observe that Regions with high texture richness are often under-optimized. Therefore, we propose \textit{TADC} dynamically control the density of Gaussians according to texture richness.
    } 
    \label{fig:tadc}
\end{figure}

\subsection{Texture-Aware Density Control}
\noindent \textbf{\textit{How to Represent Texture Richness?}} 
Regions with significant color variation
(i.e., large color gradients) contain rich texture details~\cite{hu2025texture,shi2025sparse4dgs}. To provide an estimate of texture richness, we apply a high-pass filter (Sobel in this paper) to compute the first-order finite differences of pixel intensities. Given an RGB image \(I\in\mathbb{R}^{H\times W\times3}\), we compute two gradient maps by convolving \(I\) with the standard horizontal and vertical Sobel operators, respectively:
\begin{equation}
    \begin{aligned}
T_x &= {I} \ast 
\begin{bmatrix}
-1 &  0 & 1\\
-2 &  0 & 2\\
-1 &  0 & 1
\end{bmatrix},
T_y &= {I} \ast 
\begin{bmatrix}
-1 & -2 & -1\\
 0 &  0 &  0\\
1 & 2 & 1
\end{bmatrix}.
\end{aligned}
\end{equation}

We then compute the gradient magnitude, thereby deriving the texture richness map $TR = \sqrt{(T_x)^2 + (T_y)^2}$.

\noindent \textbf{\textit{Observations.}} The proposed Gaussian prediction module predicts one Gaussian primitive for each pixel. However, we observed that regions with high texture richness are often under-optimized and show large inconsistencies with the ground truth in error maps (see Fig.~\ref{fig:tadc}). More Gaussians primitives are required to fit these texture-rich areas~\cite{hu2025texture}. To this end, we propose \textit{Texture-Aware Density Control (TADC)}, a module that dynamically controls the density of Gaussians according to texture richness.

\label{sec:TADC}
\noindent \textbf{\textit{Texture Richness Perceptron for LR Images.}} To begin with, we need to recognize where the texture-rich regions are from the 2D LR images. To achieve this, we carefully designed the \textit{Texture Richness Perceptron} $\mathcal{E}_{tex}$,  which is implemented using a convolutional neural network. Given an LR input image $I^{\text{LR}}$, we predict its texture richness map $\hat{TR}=\mathcal{E}_{tex}(I^{\text{LR}})$.
To optimize $\mathcal{E}_{tex}$, we employ the texture richness map $TR$ obtained as described above as supervisory signal:
\begin{equation}
\label{eq:TI_l1}
    \mathcal{L}_{\mathrm{tex}}
=\mathcal{L}_1\bigl(\hat{TR},TR\bigr)\,.
\end{equation}

\noindent \textbf{\textit{Density Control.}} After obtaining the predicted Gaussian set \(G_{\text{coarse}}\) and the texture richness map $\hat{TR}$, we select the Gaussian primitives set ${G}_{den}$ exhibiting the high texture richness. Each selected Gaussian primitive is further decomposed into multiple finer primitives. 

Specifically, their positions and features, denoted as \(\mathcal{X}_{\mathrm{den}}\) and \(\mathcal{F}_{\mathrm{den}}\), are input to a dedicated densification network $\mathcal{N}_{\mathrm{dens}}$ that predicts the upsampled positions and features~\cite{GenerativeDensification}. These upsampled positions and features are subsequently passed to a \emph{HEAD} module, where they are converted into fine Gaussian parameters. The set of densified Gaussian primitive is defined as:
\begin{equation}
\label{eq:dense}
G_{\mathrm{dense}} ={\textit{HEAD}}(\mathcal{N}_{\mathrm{dens}} (\mathcal{X}_{\mathrm{den}},\mathcal{F}_{\mathrm{den}}) , {G}_{den}).
\end{equation}
Finally, we obtain the refined Gaussian primitives $G_{\mathrm{refine}}$:
\begin{equation}
\label{eq:refine}
G_{\mathrm{refine}}
    = G_{\mathrm{dense}} \cup G_{\mathrm{coarse}}\bigr.
\end{equation}

\subsection{Training Loss}
During training, we directly supervise the quality of the novel RGB images using Mean Squared Error (MSE) and Learned Perceptual Image Patch Similarity (LPIPS)~\cite{zhang2018unreasonable} losses. We also introduce $\mathcal{L}_{\mathrm{tex}}$ to supervise the texture richness perceptron. The total training loss is:
\begin{equation}
\label{eq:loss}
\mathcal{L} = \lambda_{\mathrm{mse}}\,\mathcal{L}_{\mathrm{mse}} + \lambda_{\mathrm{lpips}}\,\mathcal{L}_{\mathrm{lpips}} +\lambda_{\mathrm{tex}}\,\mathcal{L}_{\mathrm{tex}} ,
\end{equation}
where $\lambda_{\mathrm{mse}}$, $\lambda_{\mathrm{lpips}}$ and $\lambda_{\mathrm{tex}}$ are balancing weights.

\begin{figure*}[t]
    \centering
    \includegraphics[width=0.92\linewidth]{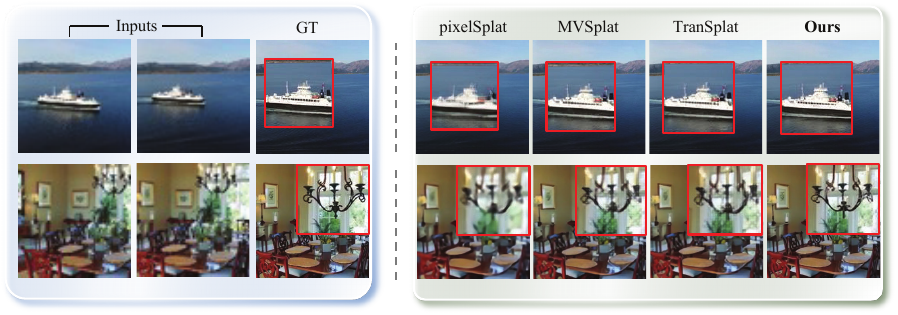}
    \caption{Qualitative comparison of novel views on RealEstate10k and ACID datasets. Compared with baseline methods, our approach yields sharper high-resolution novel views with cleaner fine-grained details (e.g., the chandelier and wall painting).
    } 
    \label{fig:duibi}
\end{figure*}

\begin{table*}[t]
    \centering 
	\scriptsize
    \begin{tabular}{c|l|ccc|ccc|ccc}
    \multirow{2}{*}{Datasets}& \multirow{2}{*}{Methods} &\multicolumn{3}{c|}{{1/2 \text{Resolution}}} & \multicolumn{3}{c|}{{1/4 \text{Resolution}}} & \multicolumn{3}{c}{{1/8 \text{Resolution}}} \\
    \cline{3-11}
     &&PSNR\(\uparrow\)&SSIM\(\uparrow\)&LPIPS\(\downarrow\)&PSNR\(\uparrow\)&SSIM\(\uparrow\)&LPIPS\(\downarrow\)&PSNR\(\uparrow\)&SSIM\(\uparrow\)&LPIPS\(\downarrow\)\\
\ChangeRT{1.2pt}     
     \multirow{4}{*}{RealEstate10K}&pixelSplat~\cite{charatan2024pixelsplat}& 24.56& 0.805  & 0.254   & \underline{22.80}& 0.720&0.358& \underline{20.64}& 0.618&0.491\\
     &MVSplat~\cite{chen2024mvsplat}& 24.06  & 0.812& 0.177& 22.20& 0.719&0.262& 20.32& 0.614&0.375\\
     &TranSplat~\cite{zhang2025transplat}& \underline{24.60}  & \underline{0.827}  & \underline{0.168}   & 22.53& \underline{0.727}&\underline{0.256}& 20.51& \underline{0.623}& \underline{0.369} \\
     &\textbf{SRSplat (Ours)}& \textbf{25.20}& \textbf{ 0.844}&  \textbf{0.152}&  \textbf{23.99}&  \textbf{0.802}& \textbf{0.189}&  \textbf{21.98}&  \textbf{0.712}& \textbf{0.267}\\
\ChangeRT{1.2pt}     
     \multirow{4}{*}{ACID}&pixelSplat~\cite{charatan2024pixelsplat}& \underline{26.56}& 0.780  & 0.284   & \underline{24.81}& \underline{0.687}&0.405& \underline{22.97}& \underline{0.603}&0.527\\
     &MVSplat~\cite{chen2024mvsplat}& 26.18  & 0.784& \underline{0.192}& 24.25& 0.675&0.282& 22.61& 0.583&\underline{0.377}\\
     &TranSplat~\cite{zhang2025transplat}& 26.25  & \underline{0.786}  & 0.193   & 24.34& 0.677&\underline{0.278}& 22.57& 0.582& 0.378 \\
     &\textbf{{SRSplat (Ours)}}& \textbf{27.39}&  \textbf{0.823}&  \textbf{0.163}&  \textbf{25.62}&  \textbf{0.757}& \textbf{0.218}&  \textbf{23.79}&  \textbf{0.660}& \textbf{0.288}\\
    \end{tabular}
    \caption{Quantitative comparison under different input resolutions. SRSplat surpass all baseline methods in terms of PSNR, SSIM, and LPIPS. (\textbf{Bold} figures indicate the best and \underline{underlined} figures indicate the second best)}
\label{tab:comparison}
\end{table*}

\begin{table}[ht]
    \centering
	\renewcommand\arraystretch{1.2}
	\scriptsize
    \begin{tabular}{l|ccc|ccc}
    \multirow{2}{*}{\textbf{{Methods}}} &\multicolumn{3}{c|}{Re10k$\rightarrow$DTU}   & \multicolumn{3}{c}{Re10k$\rightarrow$ACID}\\
    \cline{2-7}
    &PSNR&SSIM&LPIPS&PSNR&SSIM&LPIPS \\
\ChangeRT{1.2pt}     
pixelSplat & 12.66 & 0.367 & 0.577 & \underline{24.70} & \underline{0.679} & 0.416 \\
     MVSplat & \underline{13.75} & \underline{0.410} & \underline{0.511} & 24.11 & 0.677 & 0.297 \\
TranSplat  & 13.56 & 0.391 & 0.532 & 24.22 & 0.675 & \underline{0.296} \\
     Ours& \textbf{13.80} & \textbf{0.417} & \textbf{0.445} &
\textbf{25.55} & \textbf{0.754} & \textbf{0.226} \\
    \end{tabular}
        \caption{{Quantative comparisons of cross-dataset generalization.} We conduct zero-shot evaluations on the ACID and DTU datasets with models trained on RealEstate10K.}      
\label{tab:cross_dataset}
\end{table}

\section{Experiments}
\subsection{Experimental Settings}
\textbf{Datasets.} Our model is trained on the large-scale RealEstate10K~\cite{zhou2018stereo} and ACID~\cite{liu2021infinite} datasets. We evaluate our method on the RealEstate10K, ACID, and DTU datasets. We downsample the original training sets by factors of 2, 4, and 8 for training and evaluation. Following MVSplat~\cite{chen2024mvsplat}, the model is trained with two context views, and all methods are evaluated on three novel target views. For the DTU dataset, results are reported on 16 validation scenes, each with four novel views. 

\noindent\textbf{Baselines and metrics.} 
We compare SRSplat with typical methods in scene-level novel view synthesis, including pixelSplat (CVPR24)~\cite{charatan2024pixelsplat}, MVSplat (ECCV24)~\cite{chen2024mvsplat}, and TranSplat (AAAI25)~\cite{zhang2025transplat}. 
We also compare the per-scene optimization method SRGS~\cite{2024srgs} and FSGS~\cite{zhu2024fsgs} on DTU dataset.
In the following sections, we report the average PSNR, SSIM~\cite{wang2004image}, and LPIPS~\cite{zhang2018unreasonable} for all baselines.

\noindent\textbf{Implementation.} We implement SRSplat using the PyTorch framework. For each scene, we downsample images by factors of 2, 4, and 8 to create LR inputs. Training is conducted with a batch size of 10 across 5 NVIDIA RTX 4090 GPUs using the Adam~\cite{kingma2014adam} optimizer for 10k iterations, requiring approximately 2 days. 

\begin{figure}[h]
    \centering
    \includegraphics[width=0.95\linewidth]{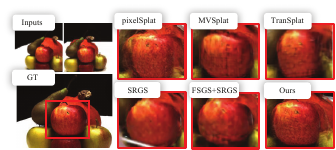}
    \caption{The qualitative comparisons on DTU datasets.
    } 
    \label{fig:dtu}
\end{figure}

\subsection{Main Results}
\noindent\textbf{Quantitative Results.} The quantitative results are presented in Tab.~\ref{tab:comparison}. SRSplat achieves state-of-the-art performance across all visual quality metrics on both RealEstate10K and ACID benchmarks. Notably, as the input image resolution decreases, the reconstruction performance of previous methods degrade significantly, while SRSplat maintains comparatively strong metrics. This is because our proposed method effectively preserves and enhances high-frequency texture representations under LR inputs.

\noindent\textbf{Qualitative Results.} Fig.~\ref{fig:duibi} presents the visualization results of SRSplat and other methods. SRSplat achieves superior quality on novel view images across various challenging scenes. In the first row (large smooth regions like the open water band), our approach reconstructs sharper object boundaries (ship hull and mast) despite scarce intrinsic texture. In texture-rich regions (second row),  our \textit{TADC} module adaptively densifies Gaussians, yielding cleaner texture details while other methods blur them. 

\begin{table}[t]
    \centering
\renewcommand\arraystretch{1.2}
\scriptsize
\begin{tabular}{l|cccc}
{\textbf{{Methods}}} 
 & PSNR$\uparrow$ & SSIM$\uparrow$ & LPIPS$\downarrow$ & Time$\downarrow$ \\
\ChangeRT{1.2pt}
SRGS~\cite{2024srgs}  & 12.42 & 0.327 & 0.598 & \underline{300s} \\
SRGS + FSGS~\cite{zhu2024fsgs}    & \underline{13.72} & \textbf{0.444} & \underline{0.481} & 420s \\
Ours        & \textbf{13.80} & \underline{0.417} & \textbf{0.445} & \textbf{0.2s} \\
\end{tabular}%
    \caption{{Quantitative results comparing with the per-scene optimization method.}}
\label{tab:SRGS_FSGS_com}
\end{table}

\begin{table}[t]
    \centering
	\renewcommand\arraystretch{1.2}
	\scriptsize
    \begin{tabular}{l|ccc|ccc}
    \multirow{2}{*}{\textbf{{Methods}}} &\multicolumn{3}{c|}{Setting(a)}   & \multicolumn{3}{c}{Setting(b)}\\
    \cline{2-7}
     &PSNR&SSIM&LPIPS&PSNR&SSIM&LPIPS\\
\ChangeRT{1.2pt}     
pixelSplat & \underline{21.38} & 0.642 & 0.415 & \underline{21.63} & \underline{0.704} & 0.365 \\
     MVSplat & {21.04} & {0.643} & 0.289 & 20.52 & 0.686 & 0.289 \\
TranSplat  & 21.24 & \underline{0.652} & \underline{0.279} & 20.88 & 0.698 & \underline{0.286} \\
     Ours& \textbf{22.07} & \textbf{0.727} & \textbf{0.219} &
\textbf{21.74} & \textbf{0.711} & \textbf{0.280} \\
    \end{tabular}
    \caption{{Quantative comparisons of cross-resolution generalization.} Setting (a): train on 1/8 resolution inputs with original-resolution supervision, test with 4× upsampling. Setting (b): train on 1/4 resolution inputs with 1/2 resolution supervision, test with 4× upsampling. }
\label{tab:cross_resolution}
\end{table}

\subsection{Other Results}
\noindent\textbf{Cross-dataset generalization.} To verify the cross-dataset generalization ability of SRSplat, we train the model on the RealEstate10K dataset downsampled by a factor of 4 and directly evaluate it on the ACID and DTU datasets. The quantitative results are presented in Tab.~\ref{tab:cross_dataset}. Compared with previous methods, SRSplat achieves significant improvements in generalization, obtaining +0.85 dB PSNR, +0.075 SSIM, and –0.07 LPIPS on the ACID dataset compared to suboptimal baselines. As depicted in Fig.~\ref{fig:dtu}, the images generated by SRSplat exhibit finer texture details and less blurriness.

\noindent\textbf{Cross-resolution generalization.} We assess cross-resolution generalization in two experimental configurations. First, we train on RealEstate10K images downsampled by a factor of 8, using original-resolution (1×) ground truth as supervision, and then perform 4× super-resolution during testing. Second, we train on inputs downsampled by a factor of 4, supervised by the 2×-downsampled ground truth, and then apply 4× super-resolution during evaluation. As shown in Tab.~\ref{tab:cross_resolution}, SRSplat exhibits superior cross-resolution generalization compared to baselines.

\noindent\textbf{Comparison with the per-scene optimization method.} We further compare SRSplat with the representative super-resolution method SRGS~\cite{2024srgs} and an enhanced version of SRGS with FSGS~\cite{zhu2024fsgs} on the DTU dataset. Qualitative results are shown in Fig.~\ref{fig:dtu} and the quantitative results are reported in Tab.~\ref{tab:cross_dataset}. Compared to per-scene optimization methods, SRSplat not only achieves competitive performance but also enables real-time reconstruction.

\begin{figure}
    \centering
    \includegraphics[width=0.9\linewidth]{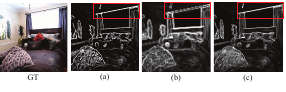}
    \caption{Comparison of texture‐richness (TR) maps. (a) TR map obtained by applying the Sobel to the HR GT image; (b) TR map extracted by Sobel on the LR image; (c) TR map perceived by our TR perceptron from the LR image.
    } 
    \label{fig:TR_ablation}
\end{figure}

\subsection{Ablation Study}
We conduct a detailed ablation study of SRSplat on the RealEstate10K datasets, downsampled by a factor of 4.

\noindent\textbf{Importance of each module of SRSplat.} In this section, we analyze the effectiveness of the proposed modules in detail. The \textit{RGFE} module compensates for the lack of high-frequency information in LR input images by leveraging external reference images. The \textit{TADC} adaptively densifies Gaussian primitives based on the texture richness of internal images. The experimental results are summarized in Tab.~\ref{tab:Ablation} (rows 1–3), demonstrating the pivotal roles of these modules in super-resolution reconstruction.

\noindent\textbf{Importance of texture richness perceptron.} The texture richness (TR) perceptron plays a crucial role in distilling texture richness from LR inputs, which is essential for subsequent texture-aware density control. When we remove TR perceptron and use the Sobel operator directly to obtain TR maps from LR images, we observe a decrease in reconstruction quality, as shown in Tab.~\ref{tab:Ablation} (row 4). Fig.~\ref{fig:TR_ablation} presents a comparison of TR maps extracted by different methods. The maps produced by the TR perceptron align more accurately with scene texture details (e.g., the curtain rod).



\begin{table}[t]
    \centering
\resizebox{0.4\textwidth}{!}{%
\renewcommand\arraystretch{1.2}
\scriptsize
\begin{tabular}{cc|ccc}
\textit{FGFE} & \textit{TADC}
 & PSNR$\uparrow$ & SSIM$\uparrow$ & LPIPS$\downarrow$ \\
\ChangeRT{1.2pt}
  &  & 22.20 & 0.719 & 0.262 \\
\checkmark & &  {23.56} & {0.782} & {0.207} \\
\checkmark & \checkmark    & \textbf{23.99} & \textbf{0.802} & \textbf{0.189} \\
\ChangeRT{1.2pt}
\multicolumn{2}{c|}{w/o TR perceptron}
&23.75 & 0.795 & 0.192
\end{tabular}%
}
    \caption{Ablation study on RealEstate10K.}
\label{tab:Ablation}
\end{table}

\section{Conclusion}

In this work, we propose \textbf{SRSplat}, the first feed-forward Super-Resolution framework. Our method initially constructs a reference gallery tailored to the unique characteristics of each scene. With this reference gallery, we introduce \textit{RGFE} to align and fuse features from the LR input images and their reference. To further refine scene details, we propose \textit{TADC}, which adaptively adjusts the density of scene’s Gaussians according to the texture richness. Extensive experiments on multiple datasets demonstrate that SRSplat outperforms existing methods.

\section*{Acknowledgments}
This work was supported in part by the National Natural Science Foundation of China under Grants (No. 62206082, 422204, 62502135), the Key
Research and Development Program of Zhejiang Province
(No. 2025C01026), the Zhejiang Provincial Natural Science Foundation of China under Grants
(No. LQN25F030014). This research was also supported by the National College Student Innovation and Entrepreneurship Training Program of China under Grants (No. 202410336018 and 202510336016).

\bibliography{main}

\end{document}